%% file: main.tex
\documentclass[10pt,letterpaper]{article}
\pdfoutput=1
\usepackage{hyperref}
\usepackage{natbib}

\usepackage{amsmath}
\usepackage{amssymb}
\usepackage{subcaption}

\usepackage{tikz}
\usetikzlibrary{shapes.misc, positioning, calc, backgrounds, decorations.pathreplacing}
\definecolor{orange}{HTML}{ff7f0e}
\definecolor{blue}{HTML}{1f77b4}

\usepackage{authblk}

\usepackage[top=1in,bottom=1in,textwidth=6 true in]{geometry}

\date{}
\title{Emergence of the Primacy Effect \\ in Structured State-Space Models}
\author{Takashi Morita}
\affil{Academy of Emerging Sciences, Chubu University}
\affil{Institute for Advanced Research, Nagoya University}
\affil{\nolinkurl{tmorita@alum.mit.edu}}

\begin{document}

\maketitle

\begin{abstract}
Structured state-space models (SSMs) have been developed to offer more persistent memory retention than traditional recurrent neural networks,
while maintaining real-time inference capabilities and addressing the time-complexity limitations of Transformers.
Despite this intended persistence, the memory mechanism of canonical SSMs is theoretically designed to decay monotonically over time, meaning that more recent inputs are expected to be retained more accurately than earlier ones.
Contrary to this theoretical expectation, however, the present study reveals a counterintuitive finding: when trained and evaluated on a synthetic, statistically balanced memorization task, SSMs 
predominantly preserve the \emph{initially} presented data in memory.
This pattern of memory bias, known as the \emph{primacy effect} in psychology, presents a non-trivial challenge to the current theoretical understanding of SSMs and opens new avenues for future research.
\end{abstract}

\begin{figure}[h]
	\centering
	\includegraphics[width=0.5\linewidth]{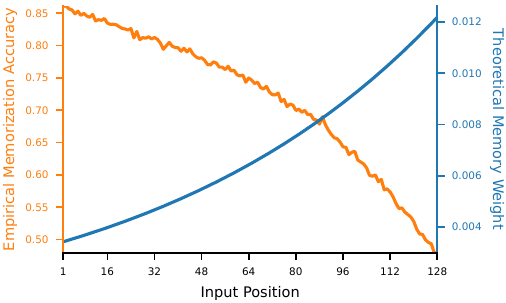}
	\caption{Graphical abstract of the key finding. When trained on a memorization task with sequentially presented items, the SSM achieves the highest accuracy for the earliest items (orange curve). This contrasts with the theoretical design of the SSM's memory mechanism, which assigns exponentially diminishing importance to older observations (blue curve).}
	\label{fig:graphical_abstract}
\end{figure}

\section{Introduction}
In recent years, \emph{structured state-space models} (SSMs) have garnered increasing attention as a backbone of next-generation artificial intelligent systems \citep{Gu+22,GuDao24}.
SSMs were developed to provide more persistent memory retention than traditional recurrent neural networks (RNNs),
while maintaining real-time inference capabilities and addressing the time-complexity limitations of Transformers.

Despite this intended persistence, the memory mechanism of canonical SSMs is theoretically designed to decay monotonically over time \citep{Gu+23_ICLR};
that is, more recent inputs are expected to be retained more accurately than earlier ones.
For example, when a sequence of random integers such as $49,75,\dots,5,38$ is presented in that order, the final items ($5,38$) are theoretically more likely to be recalled accurately at the end of the sequence (blue curve in Figure~\ref{fig:graphical_abstract}).

Contrary to this theoretical expectation, however, the present study reveals a counterintuitive finding: when trained and evaluated on a synthetic, statistically balanced memorization task, SSMs 
predominantly preserve the \emph{initially} presented data (e.g., $49,75$) in memory (orange curve in Figure~\ref{fig:graphical_abstract}).
This memory bias is known as the \emph{primacy effect} in psychology;
human and animal memory for sequentially presented items tends to be more accurate for those appearing at the beginning of the sequence \citep[][]{Ebbinghaus13,Murdock62,GlanzerCunitz66}.%
\footnote{
	Human and animal memory is also known to be more accurate for the most recently observed items---a phenomenon termed the \emph{recency effect}.
	The theoretical design of SSMs aligns with this opposite pattern.
}
This finding presents a non-trivial puzzle for existing theories and opens new avenues for research on SSMs.

The remainder of this paper is organized as follows. \textsection\ref{sec:preliminary} first reviews the formalization and memory characteristics of SSMs.
The section also discusses data-driven primacy effects observed in large language models---which inherit human biases embedded in linguistic data---and contrasts these findings with the statistically balanced setting examined in this study.
\textsection\ref{sec:methods} then details the task and model specifications, followed by results in \textsection\ref{sec:results}. Finally, \textsection\ref{sec:discussions} summarizes the findings and discusses their implications in the context of prior work.

\section{Preliminaries and Related Studies}
\label{sec:preliminary}

\subsection{Structured State-Space Models}
\subsubsection{Formalization}
\label{sec:preliminary_SSM}

To achieve more persistent memory retention than traditional RNNs,
the SSM leverages continuous-time dynamics, replacing the discrete-time framework of RNNs \citep{Zhang+18_FRU,Voelker+19,Gu+20}.
The fundamental principle underlying these models is the polynomial approximation of observed signals \citep[called the \emph{HiPPO} framework;][]{Gu+20}.
Specifically, given a \emph{single}-channel input signal $x(\cdot)$, represented as a function of time, its approximation up to a given time point $t$ can be approximated by a linear combination of polynomials:
\begin{align}
	x|_{\leq t}(\cdot)
	\approx \sum_{n=0}^{N-1} h_n(t) P_n(\cdot)
\end{align}
where $P_n$ denotes the basis polynomial of degree $n$ and $\{ h_n(t) \}_{n=0}^{N-1}$ are the optimal coefficients for the approximation at time $t$.
When these polynomials form an orthogonal basis with respect to a time-dependent measure $d\mu^{(t)}(\cdot)$, the optimal coefficients can be determined as:
\begin{align}
	h_n(t) = \langle x|_{\leq t}, P_n \rangle = \int x|_{\leq t}(s) P_n(s)\,d\mu^{(t)}(s)
\end{align}
Then, these coefficients offer a finite- and constant-dimensional representation of the input signal up to time $t$.
The framework can be naturally extended to multi-channel signals by performing channel-wise approximations in parallel, yielding $h_n^{(m)}(t)$ for each channel $m$.

For the polynomial coefficients to serve as a ``memory'' of the input signal, their temporal evolution must be trackable in an \emph{online} manner; that is, the polynomial approximation at time $t$ should not refer back to past values of the input signal, $x|_{\leq t}(s)$ ($0 \leq s <t$).
Fortunately, for certain families of polynomials, including Legendre, Laguerre, and Fourier basis,%
\footnote{
	The Fourier approximation (or transform) is not based on polynomials, but the theory can be generalized to incorporate it by taking the complex-valued basis $z^n := e^{2\pi ins}$ and a measure on the unit circle \citep{Gu+20}.
}
the coefficient dynamics can be described by an ordinary differential equation \citep[ODE;][]{Gu+20}:
\begin{align}
	\frac{d}{dt}\mathbf{h}(t) = A \mathbf{h}(t) + B x(t)
	\label{eq:ssm}
\end{align}
where $\mathbf{h}(t) := ( h_0(t), \dots, h_{N-1}(t) )^{\mathsf{T}}$,
and $A$ and $B$---referred to as the \emph{state} and \emph{input} matrices, respectively---are of size $N \times N$ and $N \times 1$.
The values on $A$ and $B$ depend on the choice of the underlying polynomial basis, and can also be adjusted via gradient-based optimization.
A feedforward transformation of the state vector $\mathbf{h}(t)$ (achieved via left-multiplication by another matrix $C$) yields a (possibly multi-channel) output signal $\mathbf{y}(t) = C \mathbf{h}(t) \in \mathbb{R}^{M}$.%
\footnote{
	The general formulation of the SSM incorporates an additional matrix $D$, which establishes a direct feedforward connection between the input and output signals, expressed as $\mathbf{y}(t) = C \mathbf{h}(t) + D \mathbf{x}(t)$.
	In practice, however, $D$ is often set to the identity matrix, effectively reducing the feedforward transformation to a simple residual connection \citep{He+16_ResNet}.
}
The resulting mapping $x \mapsto \mathbf{y}$ defines the SSM \citep[][]{Gu+21,Gu+22}.


In practice, continuous-time recordings of an input signal $x(t)$ are not available; instead, empirical data consist of discrete-time samples at $t = t_1, \dots, t_L$.
Consequently, the SSM matrices must also be discretized in order to convert the ODE in Eq.~\ref{eq:ssm} to a discrete recurrent system, analogous to RNNs:
\begin{align}
	\mathbf{h}(t_j) = \bar{A} \mathbf{h}(t_{j-1}) + \bar{B} x(t_j)
	\label{eq:ssm_disc}
\end{align}
where $\bar{A}$ and $\bar{B}$ represent the discretized versions of $A$ and $B$, respectively.
A commonly used discretization technique is the bilinear method \citep{Tustin47}, which yields:
\begin{align}
	\bar{A} :=&
		\left(I - \frac{\Delta t}{2} A\right)^{-1}
		\left(I + \frac{\Delta t}{2} A\right)
		&
	\bar{B} :=&
		\left(I - \frac{\Delta t}{2} A\right)^{-1}
		\Delta t B
\end{align}
where $\Delta t:= t_{j+1} - t_{j}$ ($\forall j = 1,\dots,L-1$) defines the time-step size.
This time-step parameter is treated as learnable, allowing the model to automatically adjust the time scale of its state-space dynamics to align with that of the input signal.%
\footnote{
	Despite the data-driven adjustability of $\Delta t$, it is important to recognize that discretization methods are generally designed under the assumption of sufficiently small values on this parameter, for effectively approximating the limit $\Delta t \to 0$.
	Consequently, setting $\Delta t$ too large introduces discrepancies between the continuous and discretized dynamics (as illustrated by the contrast between the left vs. right panels in Figure~\ref{fig:kernel}).
}
Moreover, in multi-channel settings, the model can represent multi-scale dynamics by assigning distinct $\Delta t^{(m)}$ values to each channel $m$.

This flexibility extends the applicability of SSMs to inherently discrete data lacking overt continuous dynamics \citep[e.g., text languages;][]{GuDao24,DaoGu24}; the model can jointly lean latent representations (or embeddings) of discrete inputs along with their (pseudo-)continuous dynamics via gradient-based optimization.
Furthermore, SSMs can be hierarchically stacked to construct deeper and more expressive models, where each layer processes latent signals received from lower layers.

\subsubsection{Prior Studies on the Memory Dynamics of SSMs}
\label{sec:preliminary_SSM_memory}

\begin{figure*}
	\centering
	\includegraphics[width=\linewidth]{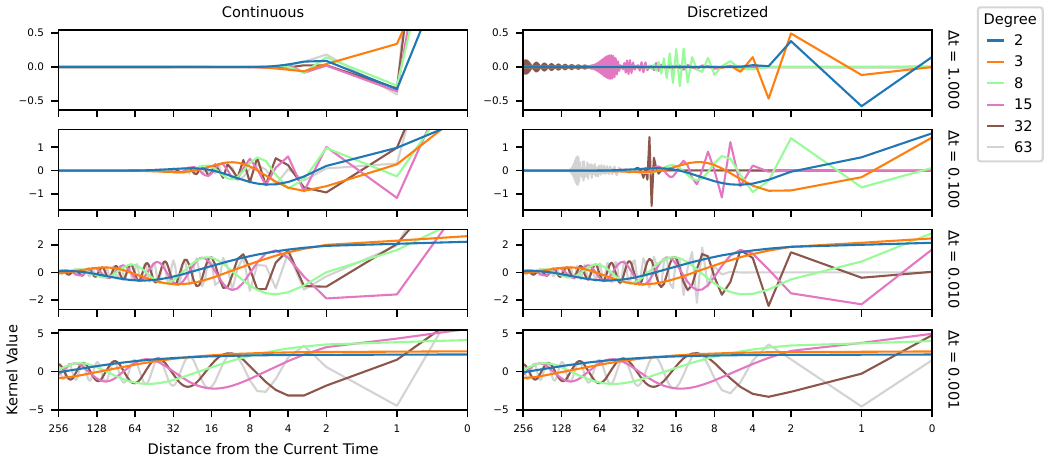}
	\caption{Illustration of the continuous (left; $K(\tau) := e^{\tau A} B$) and discretized (right; $\bar{K}_{j} := \bar{A}^{j} \bar{B}$) SSM kernels based on Legendre polynomials with an exponentially decaying measure \citep{{Gu+23_ICLR}}.
	To aid intuitive understanding, the horizontal axis has been flipped, so that kernel values multiplied with past inputs appear on the left (unlike in the standard visualization of convolutional kernels, where they are placed on the right).
	The distinct line colors represent selected entries of the kernel vectors, $K_N(\tau)$ and $\bar{K}_{j,n}$, each corresponding to the approximating polynomial of degree $n \in \{ 2, 3, 8, 15, 32, 63\}$.
	The kernels were evaluated at $\tau = j\Delta t$ for $j=0,\dots,255$ and $\Delta t \in \{0.001, 0.01, 0.1, 1.0\}$.
	Increasing $\Delta t$ results in growing discrepancies between the continuous and discretized kernels.
	}
	\label{fig:kernel}
\end{figure*}

Previous studies have identified that the time-step size, $\Delta t$, as a critical factor in determining the success/failure of SSMs.
Intuitively, a small $\Delta t$ results in minor state updates, yielding slow dynamics in the state space, $\mathbf{h}(t_{j+1})-\mathbf{h}(t_{j})$;
conversely, a large $\Delta t$ induces rapid state transitions \citep{Gu+21,GuDao24}.
As a consequence, $\Delta t$ governs the memory decay properties of the SSM;
although the models assume an exponentially decaying measure in continuous time \citep[particularly when employing Legendre/Laguerre polynomials;][]{Gu+20,Gu+23_ICLR}, choosing small $\Delta t$ values can allocate relatively large weights to input samples from distant time steps, thereby compromising single-step discriminability, which is better preserved with larger $\Delta t$ (Figure~\ref{fig:kernel}).
Notably, when an SSM employs a measure with fixed-length support---such as that used in the Fourier basis---$\Delta t^{-1}$ corresponds to the model's effective memory length \citep{Gu+23_ICLR}.

Owing to the rich theoretical foundations, previous experimental investigations into the memory capacity of SSMs have remained relatively cursory.
In particular, most prior works have only reported time-averaged benchmark scores, without offering a detailed analysis of the temporal patterns in memorized vs. overlooked information \citep{Gu+20,Gu+21,Gu+22,Gu+22_S4D,Gupta+22}.
The present study addresses this underexplored question and reveals a primacy effect in SSMs, which stands in stark contrast to their theoretically prescribed memory decay.

\subsection{Data-Driven Primacy Effect in Language Models}
\label{sec:preliminary_LLM}

Several prior studies have documented the primacy effect of artificial neural networks trained on the autoregressive language modeling task.
\citet{Wang+23_primacy-effect} investigated positional biases in a Transformer-based large language model (LLM) using a prompting-based approach.
Specifically, a list of action or event labels was sequentially presented (e.g., ``Label 1: change\_pin'', ``Label 2: card\_arrival'', ``Label 3: activate\_my\_card'').
The model was then given a query prompt specifying a target action/event (e.g., ``Target Text: I need a new PIN.'') along with an instruction statement (e.g., ``Which label matches the intent expressed in the Target Text?'').
The primacy effect was observed as a greater frequency of the initially presented labels in the model's responses.
Comparable findings have been reported across different LLM implementations and benchmark datasets \citep{EicherIrgolic24,GuoVosoughi24,Janik24,Liu+24}.

\citet{Xiao+24} found that Transformer-based LLMs allocated disproportionate attention to initial tokens, regardless of their informational salience in the text (a phenomenon they termed \emph{attention sinks}).
Furthermore, retaining these initial tokens even after they fall outside the predefined input window was found to enhance model performance.

Since the Transformer architecture is inherently position-agnostic---lacking an intrinsic ordering mechanism apart from external positional encodings \citep{Vaswani+17_AttentionIsAllYouNeed}---the primacy effects observed in these studies must stem from the statistical properties of the training data or task design.
\citeauthor{Wang+23_primacy-effect} argued that LLMs inherit cognitive biases from human-generated linguistic data.
\citeauthor{Xiao+24} suggested that the nature of the language modeling task itself encourages prioritization of initial tokens, as they are repeatedly used as inputs for autoregressive predictions, reinforcing attention allocation to them.

In contrast to these prior investigations, the present study examines the emergence of the primacy effect in the SSM while \emph{preventing the inheritance of human-induced bias}.
Specifically, the models are trained on a synthetic memorization task designed based on psychological experiments conducted with humans and other animals \citep{ThompsonHerman77,SandsWright80,Wright+85}.
The following section details the task formulation and the model architecture.

\section{Methods}
\label{sec:methods}

\subsection{Task}
\label{sec:task}

\begin{figure*}
	\centering
	\scalebox{0.87}{
	\input{figs/binary-verification.tex}
	}
	\caption{Schematic illustration of the binary memory verification task.}
	\label{fig:task}
\end{figure*}

The memorization patterns of the SSM were assessed using the binary memory verification task \citep[Figure~\ref{fig:task}; a.k.a. \emph{serial probe recognition} in psychology and ethology;][]{WickelgrenNorman66,ThompsonHerman77,SandsWright80,Wright+85}.%
\footnote{
	The most widely adopted task for assessing the primacy effect in human memory is \emph{free recall}, in which participants are presented with a sequence of items and subsequently asked to recall them in an order-agnostic manner \citep{Murdock62,GlanzerCunitz66}. 
	While this paradigm can be technically formulated as a loss function---under the framework of the optimal transport \citep{Cuturi13}---initial explorations of this study revealed that model performance remained suboptimal under this approach, yielding lower accuracy than in theoretically more demanding tasks requiring order-sensitive reconstruction.
	Consequently, the present study adopted the more machine learning-friendly task based on binary verification.
	Remarkably, this task has also been used to assess the memory capacity of non-human animals, which are unable to perform free recall \citep{ThompsonHerman77,SandsWright80,Wright+85}.
}
In this task, the models were first presented with a sequence of randomly generated, non-repeating integers (hereinafter referred to as \emph{study items}).
Subsequently, they received another sequence of integer queries and were trained to determine whether each query token was present (labeled as 1) or absent (labeled as 0) in the study items.
To construct these queries, the study items were first shuffled, and then, with a probability of $p=0.5$, each shuffled token was replaced with a randomly sampled integer from the complement set of the study items (termed \emph{distractors}).%
\footnote{
	An anonymous reviewer noted that the adopted memorization task introduces a non-uniform distribution of relative distances between study items and queries.
	Specifically, the model has a lower probability of encountering distances of length $L \pm \alpha$, where $L$ is the number of study items, as $\alpha$ increases from $0$ to $L-1$.
	Nevertheless, the task preserves \emph{symmetry} between short and long distances, since distances of length $L-\alpha$ and $L+\alpha$ occur with equal frequency.
	Therefore, the statistical properties of the task do not inherently favor the memorization of earlier study items (i.e., long input-output dependencies).
	\label{ft:distance-frequency}
}

The task hyperparameters were manually adjusted to prevent the models from achieving perfect accuracy.
Specifically, the input length was set to $L \in \{64, 128, 256\}$, and the vocabulary size was fixed at $K:=4096$.
Each model underwent ten independent training runs with different random seeds.
For evaluation, 1024 sets of integers were held out as test data, ensuring that these integer combinations never appeared as study items in the training set, regardless of their order.

To build test sequences, the held-out study items were randomly ordered, and queries were generated by first shuffling and then cyclically shifting them (e.g., $(2,8,11,29)\allowbreak\mapsto\allowbreak\{ (2,8,11,29),\allowbreak(8,11,29,2),\allowbreak(11,29,2,8),\allowbreak(29,2,8,11) \}$).
This design ensured that each study item was queried in all $L$ possible positions.
Finally, either the even- or odd-indexed query positions were replaced with random distractors, resulting in a total of $1024 \times L \times 2$ test sequences per trial.

In Appendix~\ref{sec:associative-recall}, the primacy effect is examined in a more advanced task---\emph{associative recall}---which has been established as a useful benchmark for evaluating the performance of language model architectures
\citep{Olsson+22,Fu+23,GuDao24}.

\subsection{Models}
\label{sec:models}

The models used for the binary memory verification task comprised three layers, as illustrated in Figure~\ref{fig:task}.
In the first layer, the input integers were embedded into 256-dimensional real-valued vectors.
These embeddings were shared between study items and query tokens.
The resulting sequence of vectors was then processed by the SSM, whose outputs were linearly projected onto binary logits to determine whether each query token was present in the study items.

This study primarily examined the single-layer S4 model as the goldstandard implementation of the SSM \citep{Gu+22}.%
\footnote{
	Recent studies have shown that the state matrix ($A$) of S4 can be simplified into a purely diagonal form without compromising performance \citep[S4D;][]{Gu+22_S4D}.
	By contrast, the original S4 model introduced an additional low-rank component to the diagonal structure (referred to as the Diagonal Plus Low Rank form, or DPLR) to ensure a mathematically well-founded state matrix.
	Notably, the diagonal variant exhibited a qualitatively similar primacy effect to the DPLR model.
	Due to the page limitations, results for the diagonal model are omitted from this paper, and all reported findings are based on the DPLR model.}
The model encoded the channel-wise dynamics of the input embeddings in a complex-valued space, with its outputs subsequently projected back into the real domain by discarding imaginary components.
The state and input matrices ($A$ and $B$ in Eq.~\ref{eq:ssm}) were initialized to approximate each channel's trajectory using Legendre/Laguerre polynomials of degrees 0--63 (HiPPO-LegS/LagT)
or a Fourier basis $\{ s_0,c_0,\dots,s_{31},c_{31} \}$, where $s_n(t) := \sqrt{2}\sin(2\pi n t)$ and $c_n(t) := \sqrt{2}\cos(2\pi n t)$ \citep[HiPPO-Fout, Fourier Recurrent Unit;][]{Zhang+18_FRU,Gu+20,Gu+23_ICLR}.
The matrices were discretized by the bilinear method \citep{Tustin47}.

For comparison, a single-layer long short-term memory (LSTM) network was also evaluated \citep[][]{HochreiterSchmidhuber97_LSTM}.
The dimensionality of both hidden and cell states was set to 256.


The models were trained for 300,000 iterations using the Adam optimizer with parameters $(\beta_0,\beta_1) := (0.9,0.99)$ \citep{KingmaBa15_Adam}.
Batch size was set to 512.
The learning rate was linearly increased from 0.0 to 0.001 over the first 1,000 iterations (\emph{warmups}) and subsequently decayed according to the cosine annealing schedule \citep{LoshchilovHutter17}.
To prevent gradient explosion, the gradient norm was clipped at 1.0.
The Python code for the experiments is available at \url{https://github.com/tkc-morita/primacy-effect.git}.

\section{Results}
\label{sec:results}

\subsection{Emergence of the Primacy Effect}
\label{sec:results_main}

\begin{figure*}
	\begin{subcaptiongroup}
		\phantomcaption\label{fig:trained_nplr}
		\phantomcaption\label{fig:frozen_nplr}
		\phantomcaption\label{fig:frozen_lagt}
		\phantomcaption\label{fig:frozen_fout}
		\phantomcaption\label{fig:frozen_dt-1.0}
		\phantomcaption\label{fig:frozen_short}
		\phantomcaption\label{fig:frozen_long}
		\phantomcaption\label{fig:lstm}
	\end{subcaptiongroup}
	\centering
	\includegraphics[width=\linewidth]{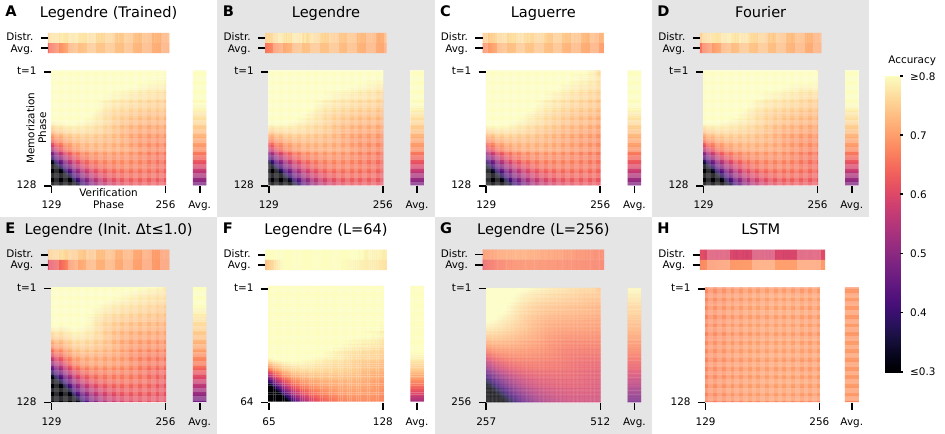}
	\caption{Accuracy of the binary memory verification task.
	Each cell in the square heatmaps represents the accuracy (or the recall score) for study items that were presented at the time indexed by the corresponding row and queried at the time indexed by the corresponding column. The accuracy for distractor queries is displayed in the top separate row of each panel, alongside the average accuracy across memorization times (rows). Similarly, the rightmost separate column represents the average accuracy across verification times (columns).
	The bottom-right panel (\subref{fig:lstm}) depicts the accuracy distribution for the LSTM, while the other panels (\subref{fig:trained_nplr}--\subref{fig:frozen_long}) report results for the SSM (S4) under different parameter configurations.
	The state and input matrices of the SSM were initialized to approximate the latent dynamics of input sequences using Legendre polynomials, except in panels \subref{fig:frozen_lagt} and \subref{fig:frozen_fout}, where Laguerre and Fourier bases were used, respectively.
	The state and input matrices were optimized for the task in panel \subref{fig:trained_nplr}, whereas they remained fixed at their initial values in all other panels.
	The discretization step size $\Delta t$ was 
	initialized in the range $0.001 \leq \Delta t \leq 0.1$, except in panel \subref{fig:frozen_dt-1.0}, where the upper bound was extended to $1.0$ (i.e., $0.001 \leq \Delta t \leq 1.0$).
	The length of study items was set to $L=128$, except in panel \subref{fig:frozen_short} ($L=64$) and panel \subref{fig:frozen_long} ($L=256$).
	}
	\label{fig:accuracy}
\end{figure*}

Figure~\ref{fig:accuracy} reports the accuracy of the binary memory verification task across all combinations of memorization and verification times.
The brightness of each cell in the square heatmaps indicates the accuracy for study items that were presented at the time indexed by the corresponding row and queried at the time indexed by the corresponding column.
That is, they report the proportion of true positives against false negatives (i.e., the \emph{recall} score).
Additionally, the top separate row of each panel displays the accuracy for distractor queries (integers not included among the study items), capturing the prevalence of true negatives over false positives.
Just below it, the second row summarizes the average accuracy across memorization times (rows).
Similarly, the rightmost separate column represents the average accuracy across all verification times (columns).

The binary memory verification performance of the SSM model was highest for study items presented at the beginning of the sequence, demonstrating a clear primacy effect (Figure~\ref{fig:trained_nplr}--\subref{fig:frozen_long}).
The model maintained high accuracy across different query timings (as indicated by the bright colors in the top rows of the heatmaps), provided that the sequence length did not exceed its capacity (see the accuracy decline in Figure~\ref{fig:frozen_long}, where $L=256$).
In other words, memory for the initial study items exhibited minimal decay over time.

By contrast, the LSTM did not display this primacy effect; its accuracy was uniform across both the memorization and verification phases (Figure~\ref{fig:lstm}).

Interestingly, the SSM's accuracy for the most recently presented study items was lowest when they were queried immediately after their initial presentation in the memorization phase (indicated by the dark colors in the bottom-left region of the heatmaps).
This suggests a temporal delay between the encoding of study items and their effective retrieval.

These findings held true regardless of whether the state and input matrices of the SSM ($A$ and $B$ in Eq.~\ref{eq:ssm}) were optimized for the task (Figure~\ref{fig:trained_nplr}) or remained fixed at their initial values (Figure~\ref{fig:frozen_nplr}--\subref{fig:frozen_long}).
Moreover, the results remained consistent across different polynomial bases underlying the state and input matrices, including Laguerre (Figure~\ref{fig:frozen_lagt}), Fourier (\ref{fig:frozen_fout}), and Legendre (all other panels).


\subsection{Distribution of the Time-Step Sizes}
\label{sec:result_dt}

\begin{figure*}
	\centering
	\includegraphics[width=1.0\linewidth]{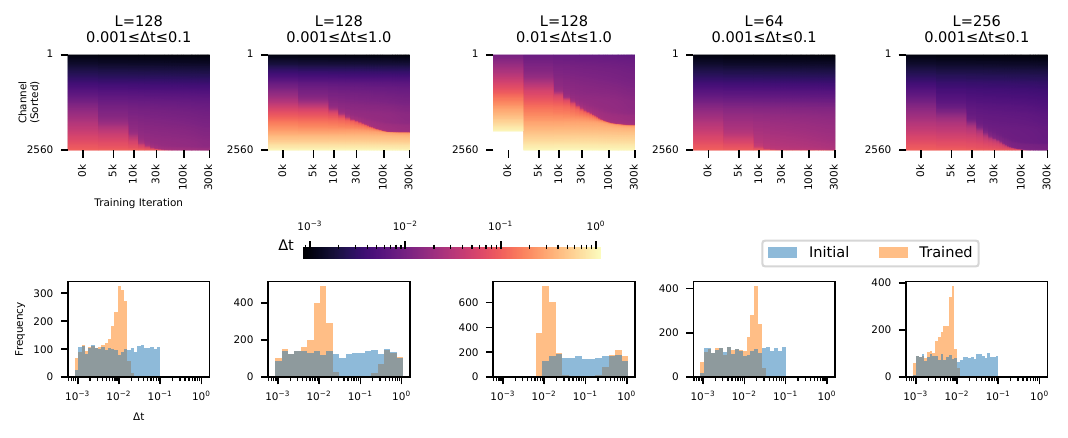}
	\caption{Top row: Optimization trajectories of the discretization step size $\Delta t$ in the SSM (S4) with frozen Legendre state and input matrices. Each heatmap column represents the distribution of $\Delta t$ across 256 latent channels $\times$ 10 training runs, sorted in the ascending order.
	Bottom row: Histograms displaying the initial (blue) and final (orange) values of $\Delta t$, aggregated across the 256 latent channels $\times$ 10 training runs (see Figure~\ref{fig:delta-t_per-seed} in Appendix~\ref{sec:delta-t_acorss_runs} for variations among individual runs).
	The first to third panel columns from the left show the results for three different ranges of log-uniformly random initializations, whereas the fourth and fifth columns tested shorter and longer study items, respectively.
	}
	\label{fig:delta-t}
\end{figure*}

As discussed in \textsection\ref{sec:preliminary_SSM_memory}, the discretization time-step size $\Delta t$ plays a critical role in determining the memory capacity of the SSM.
Moreover, once the state matrix $A$ is fixed, $\Delta t$ becomes the sole parameter capable of influencing the \emph{dynamics} of the SSM;%
\footnote{
	Freezing $\Delta t$ resulted in a complete failure of learning.}
all remaining parameters are confined to \emph{feedforward} transforms.

Accordingly, to further investigate its role, the optimization trajectories of $\Delta t$ were tracked over the course of training.
The analysis revealed that as training progressed, a specific range of step sizes ($\Delta t \leq 0.03$) became dominant, compensating for the deallocation of the higher range approximately between 0.03 and 0.2 (Figure~\ref{fig:delta-t}).

Additionally, the peak value of $\Delta t$ was found to depend on the number of the study items $L$; 
longer study sequences led the model to favor smaller $\Delta t$ values (compare the leftmost panel with the two rightmost panels).

\section{Discussions}
\label{sec:discussions}


The present study demonstrated that the SSM exhibits the primacy effect in memorization.
When performing the binary memory verification task, which parallels paradigms used to investigate memory capacity in humans and animals,
the model showed the highest accuracy for study items presented at the beginning of the sequence.
Moreover, memorized information was not retrievable immediately after the presentation of the study items.
These findings are novel and counterintuitive, as they challenge the theoretical formulation of the SSM, which assumes an exponentially decaying measure \citep[for Legendre and Laguerre bases; Fig~\ref{fig:kernel};][]{Gu+20,Gu+23_ICLR}.

As noted in \textsection\ref{sec:preliminary_SSM}, the SSM was designed to achieve a longer-lasting memory than classical RNNs \citep{Gu+20}.
Prior research on RNNs and SSMs has focused on their ability to preserve input data against temporal decay.
%
%
However, little attention has been given to how the model handles longer study sequences and larger vocabularies when memory capacity reaches its limit.%
\footnote{
	It should be noted that the performance of the SSM can be enhanced by increasing the number of layers and/or latent channels.
	In this study, the model's capacity was intentionally constrained in order to study its behavior under conditions where perfect accuracy is unattainable.
}
In particular, the question of whether the models prioritize initial/middle/recent observations has remained unexplored.
The present study addressed this question and discovered that the SSM predominantly preserved the initial observations.

The key factor responsible for the primacy effect in the SSM appears to be the time-step size, $\Delta t$, as all the other trainable parameters pertain exclusively to feedforward transforms.
After training on the memorization task, $\Delta t$ values concentrated below a specific threshold ($\Delta t \leq 0.03$).
As discussed in \textsection\ref{sec:preliminary_SSM_memory}, smaller $\Delta t$ values allow the model to retain more distant memories, while larger $\Delta t$ values enhance the discrimination of adjacent tokens \citep[][]{Gu+21,GuDao24}.
The learning results thus align with the \emph{necessary} condition for the primacy effect;
however, the question remains open why recent observations were remembered less accurately despite the exponentially decaying measure underlying the polynomial-approximation theory.
Future research may address this issue through comparisons across a wider range of discretization methods---such as the Runge-Kutta method---extending beyond the standard empirical options of bilinear and zero-order hold.

The SSMs analyzed in this study were trained from scratch on a synthetic memorization task that was designed to closely resemble controlled psychological experiments \citep{WickelgrenNorman66,ThompsonHerman77,SandsWright80,Wright+85}.
Consequently, the observed primacy effect is attributed to the intrinsic properties of the SSM per se, rather than to biases introduced by data or task design.
From this perspective, the the present study stands in contrast to prior investigations of LLMs \citep{Wang+23_primacy-effect,EicherIrgolic24,GuoVosoughi24,Janik24,Liu+24,Xiao+24};
LLMs are trained on human-generated linguistic data and therefore likely to inherit the primacy effect as a byproduct of human cognitive biases embedded in the data.

It also remains an open question whether the primacy effect holds in more advanced settings than those examined in this study.
Specifically, the scope was restricted to single-layered models, whereas empirical applications almost invariably employ multi-layered architectures.
Such extended architectures achieved perfect accuracy on the adopted task---even at the maximal levels of input length and vocabulary size implementable within the available computational resources---thereby failing to incur the memory load necessary for evaluating the primacy effect.
For the same reason, the proposed experimental paradigm was also inadequate for testing the SSM-based language model, Mamba \citep{GuDao24,DaoGu24}.
Therefore, 
Clarifying whether the primacy effect persists in such powerful architectures is an important avenue for future research.

\section*{Acknowledgments}
\lccode`\0`\0
\lccode`\1`\1
\lccode`\2`\2
\lccode`\3`\3
\lccode`\4`\4
\lccode`\5`\5
\lccode`\6`\6
\lccode`\7`\7
\lccode`\8`\8
\lccode`\9`\9
\hyphenation{JP-24-H-0-0-7-7-4 JP-22-H-0-3-9-1-4 JP-24-K-1-5-0-8-7 JP-MJCR-25-U6 JP-MJCR-22-P5 K35-XXVIII-620 JP-MJAX-21-AN}
This study was supported by 
JST AIP Accelerated Program (JPMJCR25U6), ACT-X (JPMJAX21AN), and Core Research for Evolutional Science and Technology (JPMJCR22P5);
JSPS Grant-in-Aid for Early-Career Scientists (JP21K17805) 
and for Scientific Research
A (JP24H00774),
B (JP22H03914),
and C (JP24K15087);
and Kayamori Foundation of Informational Science Advancement (K35XXVIII620).
The author also gratefully acknowledges the support of the
Academic Center for Computing and Media Studies, Kyoto University,
regarding the use of their supercomputer system.



\bibliographystyle{apalike}
\bibliography{takashi_references}

\appendix

\section{Distributions of $\Delta t$ across Training Runs}
\label{sec:delta-t_acorss_runs}

\begin{figure*}[h]
	\centering
	\includegraphics[width=1.0\linewidth]{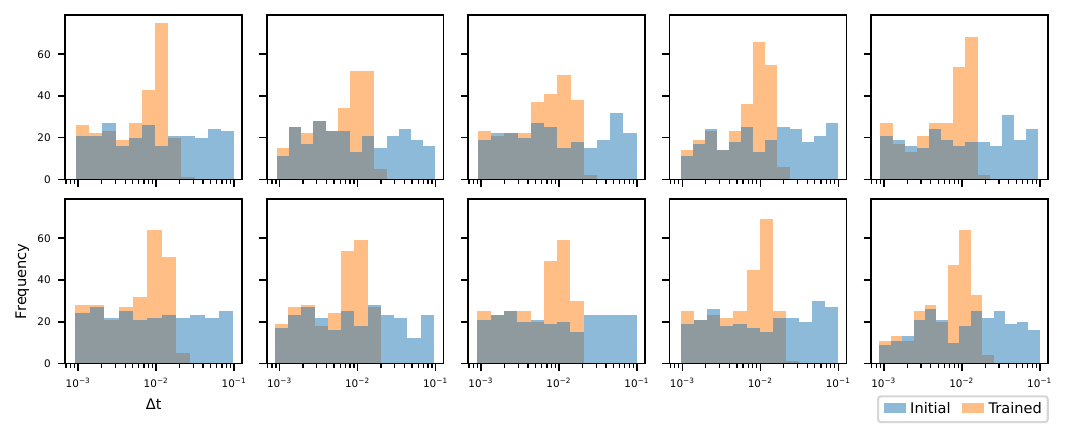}
	\caption{Histograms displaying the initial (blue) and final (orange) values of $\Delta t$. Each panel presents the distributions obtained from one of the ten training runs with different random seeds.}
	\label{fig:delta-t_per-seed}
\end{figure*}

Figure~\ref{fig:delta-t_per-seed} reports the distribution of the $\Delta t$ parameters before and after each of the ten training runs (decomposing the leftmost panel in Figure~\ref{fig:delta-t}).

\section{Associative Recall}
\label{sec:associative-recall}

This section investigates the primacy effect in a more advanced task---\emph{associative recall} \citep{Fu+23}---compared to the binary memory verification.
Similar to the memory verification setup, the associative recall task first presents a sequence of random tokens (study items) to the model, followed by a shuffled version of the same sequence (excluding the final token; Figure~\ref{fig:task_associative-recall}).%
\footnote{
	In the original formulation of the associative recall task, the model is presented with a single query token sampled from the study items. This study extends the task in a more empirical setting, requiring the model to recall multiple pairs of precedent and successor tokens within a single sequence.
}
Unlike the verification task, however, none of the shuffled query tokens are replaced with distractors; instead, the model is required to return the \emph{immediate successor} of each query token.
For example, given study items $(8,29,2,17)$ and queries $(2,8,29)$, the correct outputs are $(17,29,2)$.

\begin{figure*}
	\centering
	\scalebox{0.87}{
	\input{figs/associative-recall.tex}
	}
	\caption{Schematic illustration of the (extended) associative recall task.}
	\label{fig:task_associative-recall}
\end{figure*}

Despite its apparent simplicity, previous work has shown that the ability to recall the successor of previously presented input tokens is a meaningful indicator of language models' capacity \citep{Olsson+22,Fu+23,GuDao24}.
As such, investigating associative recall helps bridge the primacy effect observed in SSMs with real-world applications, particularly in language modeling.

Model and task hyperparameters were manually tuned to ensure that overall accuracy remained suboptimal.
Specifically, the length of the study items was set to $L=96$, and the vocabulary size to $K=128$.
The number of latent channels in the SSM (also equal to the dimensionality of the input embeddings) was set to 1024.
All other hyperparameters were identical to those used in the memory verification task (see \textsection\ref{sec:task}).

\begin{figure*}
	\begin{subcaptiongroup}
		\phantomcaption\label{fig:trained_nplr_associative-recall}
		\phantomcaption\label{fig:frozen_nplr_associative-recall}
		\phantomcaption\label{fig:frozen_lagt_associative-recall}
		\phantomcaption\label{fig:frozen_fout_associative-recall}
	\end{subcaptiongroup}
	\centering
	\includegraphics[width=1.0\linewidth]{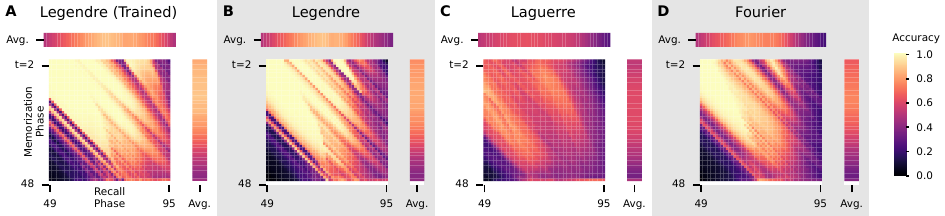}
	\caption{Accuracy of the associative recall task performed by the SSM (S4) under different parameter configurations.
	Each cell in the square heatmaps represents the accuracy (or the recall score) for study items (target successors) that were presented at the time indexed by the corresponding row and queried at the time indexed by the corresponding column.
	The top separate row and the right separate column display the average accuracy across memorization times (rows) and recall times (columns), respectively.
	The state and input matrices of the SSM were initialized to approximate the latent dynamics of input sequences using the Legendre (\subref{fig:trained_nplr_associative-recall},\subref{fig:frozen_nplr_associative-recall}), Laguerre (\subref{fig:frozen_lagt}), and Fourier (\subref{fig:frozen_fout}) polynomials.
	The state and input matrices were optimized for the task in panel \subref{fig:trained_nplr}, and remained fixed at their initial values elsewhere.
	The discretization step size $\Delta t$ was 
	initialized in the range $0.001 \leq \Delta t \leq 0.1$.
	The length of study items was set to $L=96$, and the vocabulary size to $K=128$.}
	\label{fig:associative-recall}
\end{figure*}

Figure~\ref{fig:associative-recall} shows the SSM's accuracy on the associative recall task.
The model exhibited higher accuracy for initial study items compared to terminal ones (vertical differences),
despite overall retention being less persistent than in the memory verification task (horizontal declines).
Besides this general trend, the model also appears to have learned delayed reconstructions of input sequences, as evidenced by the diagonal patterns observed in the heatmaps.
This behavior may reflect a statistical property of the adopted task;
input-output distances of length $L$ occurred more frequently than either shorter or longer dependencies during training (see Footnote~\ref{ft:distance-frequency}).

\end{document}

%% file: figs/binary-verification.tex
\begin{tikzpicture}[node distance=1.4em,font=\fontsize{8pt}{8pt}\selectfont]
    \node[rounded rectangle,draw,align=center] (RNN1) at (0,0) {SSM};
    \node[rounded rectangle,draw,right=of RNN1,align=center] (RNN2) {SSM};
    \node[rounded rectangle,draw,right=of RNN2,align=center] (RNN3) {SSM};
    \node[right=of RNN3] (RNN_dots) {$\cdots$};
    \node[rounded rectangle,draw,right=of RNN_dots,align=center] (RNN4) {SSM};
    \node[rounded rectangle,draw,right=of RNN4,align=center] (RNN5) {SSM};
    \node[rounded rectangle,draw,right=of RNN5,align=center] (RNN6) {SSM};
    \node[rounded rectangle,draw,right=of RNN6,align=center] (RNN7) {SSM};
    \node[right=of RNN7] (RNN_dots2) {$\cdots$};
    \node[rounded rectangle,draw,right=of RNN_dots2,align=center] (RNN8) {SSM};
    \draw[->,ultra thick] (RNN1) -- (RNN2);
    \draw[->,ultra thick] (RNN2) -- (RNN3);
    \draw[->,ultra thick] (RNN3) -- (RNN_dots);
    \draw[->,ultra thick] (RNN_dots) -- (RNN4);
    \draw[->,ultra thick] (RNN4) -- (RNN5);
    \draw[->,ultra thick] (RNN7) -- (RNN_dots2);
    \draw[->,ultra thick] (RNN_dots2) -- (RNN8);
    \draw[->,ultra thick] (RNN6) -- (RNN7);
    \draw[->,ultra thick] (RNN5) -- (RNN6);
    \path let \p1=(RNN1.west), \p2=(RNN8.east) in (RNN1) to (RNN8)
        node[anchor=west, minimum width=\x2-\x1+0.5em, rounded rectangle, draw,yshift=-2em] (embed_layer)
        at (RNN1.west|-RNN8.south)
        {Embedding};
    \draw[->,ultra thick,violet] (RNN1|-embed_layer.north) -- (RNN1);
    \draw[->,ultra thick,violet] (RNN2|-embed_layer.north) -- (RNN2);
    \draw[->,ultra thick,violet] (RNN3|-embed_layer.north) -- (RNN3);
    \draw[->,ultra thick,violet] (RNN4|-embed_layer.north) -- (RNN4);
    \draw[->,ultra thick,blue] (RNN5|-embed_layer.north) -- (RNN5);
    \draw[->,ultra thick,orange] (RNN6|-embed_layer.north) -- (RNN6);
    \draw[->,ultra thick,blue] (RNN7|-embed_layer.north) -- (RNN7);
    \draw[->,ultra thick,orange] (RNN8|-embed_layer.north) -- (RNN8);
    \node[yshift=-2em,text=violet] (input1) at (RNN1|-embed_layer.south) {8};
    \node[text=violet] (input2) at (RNN2|-input1) {29};
    \node[text=violet] (input3) at (RNN3|-input1) {2};
    \node[text=violet] (input_dots) at (RNN_dots|-input1) {$\cdots$};
    \node[text=violet] (input4) at (RNN4|-input1) {11};
    \node[text=blue] (input5) at (RNN5|-input1) {2};
    \node[text=blue] (input_dots2) at (RNN_dots2|-input1) {$\cdots$};
    \node[text=blue,shape=cross out,draw=orange,thick] (input6) at (RNN6|-input1) {8};
    \node[text=orange,anchor=north] (input6other) at (input6.south) {17};
    \node[text=blue] (input7) at (RNN7|-input1) {11};
    \node[text=blue,shape=cross out,draw=orange,thick] (input8) at (RNN8|-input1) {29};
    \node[text=orange,anchor=north] (input8other) at (input8.south) {34};
    \draw[->,ultra thick,violet] (input1) -- (RNN1|-embed_layer.south);
    \draw[->,ultra thick,violet] (input2) -- (RNN2|-embed_layer.south);
    \draw[->,ultra thick,violet] (input3) -- (RNN3|-embed_layer.south);
    \draw[->,ultra thick,violet] (input4) -- (RNN4|-embed_layer.south);
    \draw[->,ultra thick,blue] (input5) -- (RNN5|-embed_layer.south);
    \draw[->,ultra thick,orange] (input6) -- (RNN6|-embed_layer.south);
    \draw[->,ultra thick,blue] (input7) -- (RNN7|-embed_layer.south);
    \draw[->,ultra thick,orange] (input8) -- (RNN8|-embed_layer.south);
    \node[anchor=north west,align=center,text=violet] (input_ann) at (input1.south west) {\textbf{Study Items}\\(Non-Repeating Random Integers)};
    \node[anchor=north,align=center,text=orange,yshift=0.5em] (other) at (input6other.south-|input_dots2) {\textbf{Replacement w/ Distractors}\\(Random Non-Repeating Complements)};
    \path let \p1=(RNN5.west), \p2=(RNN8.east) in (RNN5) to (RNN8)
        node[above=2em of RNN5.north west, anchor=west, minimum width=\x2-\x1+0.5em, rounded rectangle, draw] (readout_layer) {Linear$\to$Sigmoid};
    \draw[->,ultra thick,blue] (RNN5) -- (RNN5|-readout_layer.south);
    \draw[->,ultra thick,orange] (RNN6) -- (RNN6|-readout_layer.south);
    \draw[->,ultra thick,blue] (RNN7) -- (RNN7|-readout_layer.south);
    \draw[->,ultra thick,orange] (RNN8) -- (RNN8|-readout_layer.south);
    \node[text=blue,yshift=2em] (sort5) at (RNN5|-readout_layer.north) {1};
    \node (sort_dots) at (RNN_dots2|-sort5) {$\cdots$};
    \node[text=orange] (sort6) at (RNN6|-sort5) {0};
    \node[text=blue] (sort7) at (RNN7|-sort5) {1};
    \node[text=orange] (sort8) at (RNN8|-sort5) {0};
    \draw[->,ultra thick,blue] (RNN5|-readout_layer.north) -- (sort5);
    \draw[->,ultra thick,orange] (RNN6|-readout_layer.north) -- (sort6);
    \draw[->,ultra thick,blue] (RNN7|-readout_layer.north) -- (sort7);
    \draw[->,ultra thick,orange] (RNN8|-readout_layer.north) -- (sort8);
    \node[anchor=south] (sort_ann) at (sort7.north) {\textbf{{\color{blue}{In\,Study\,Items\,(1)}} OR \color{orange}{Not\,(0)}}};
    \path let \p1=(input1.west), \p2=(input4.east) in (input1) to (input4)
        node[anchor=west, minimum width=\x2-\x1+1em, rounded rectangle, draw, dotted,thick,blue] (inputs) at (input1.west|-input4.east) {\phantom{M}};
    \path let \p1=(input5.west), \p2=(input8.east) in (input5) to (input8)
        node[anchor=west, minimum width=\x2-\x1+1em, rounded rectangle, draw, dotted,thick,blue] (shuffled) at (input5.west|-input8.east) {\phantom{M}};
    \draw[->,ultra thick,blue] (input4.south west) to [bend right=45] node [below,anchor=north west] {\textbf{Shuffle}} (input5.south east);
    \end{tikzpicture}

%% file: figs/associative-recall.tex
\begin{tikzpicture}[node distance=1.4em,font=\fontsize{8pt}{8pt}\selectfont]
    \node[rounded rectangle,draw,align=center] (RNN1) at (0,0) {SSM};
    \node[rounded rectangle,draw,right=of RNN1,align=center] (RNN2) {SSM};
    \node[rounded rectangle,draw,right=of RNN2,align=center] (RNN3) {SSM};
    \node[rounded rectangle,draw,right=of RNN3,align=center] (RNN3_2) {SSM};
    \node[right=of RNN3_2] (RNN_dots) {$\cdots$};
    \node[rounded rectangle,draw,right=of RNN_dots,align=center] (RNN4) {SSM};
    \node[rounded rectangle,draw,right=of RNN4,align=center] (RNN4_2) {SSM};
    \node[rounded rectangle,draw,right=of RNN4_2,align=center] (RNN5) {SSM};
    \node[rounded rectangle,draw,right=of RNN5,align=center] (RNN6) {SSM};
    \node[rounded rectangle,draw,right=of RNN6,align=center] (RNN7) {SSM};
    \node[right=of RNN7] (RNN_dots2) {$\cdots$};
    \node[rounded rectangle,draw,right=of RNN_dots2,align=center] (RNN8) {SSM};
    \draw[->,ultra thick] (RNN1) -- (RNN2);
    \draw[->,ultra thick] (RNN2) -- (RNN3);
    \draw[->,ultra thick] (RNN3) -- (RNN3_2);
    \draw[->,ultra thick] (RNN3_2) -- (RNN_dots);
    \draw[->,ultra thick] (RNN_dots) -- (RNN4);
    \draw[->,ultra thick] (RNN4_2) -- (RNN5);
    \draw[->,ultra thick] (RNN4) -- (RNN4_2);
    \draw[->,ultra thick] (RNN7) -- (RNN_dots2);
    \draw[->,ultra thick] (RNN_dots2) -- (RNN8);
    \draw[->,ultra thick] (RNN6) -- (RNN7);
    \draw[->,ultra thick] (RNN5) -- (RNN6);
    \path let \p1=(RNN1.west), \p2=(RNN8.east) in (RNN1) to (RNN8)
        node[anchor=west, minimum width=\x2-\x1+0.5em, rounded rectangle, draw,yshift=-2em] (embed_layer)
        at (RNN1.west|-RNN8.south)
        {Embedding};
    \draw[->,ultra thick,violet] (RNN1|-embed_layer.north) -- (RNN1);
    \draw[->,ultra thick,violet] (RNN2|-embed_layer.north) -- (RNN2);
    \draw[->,ultra thick,violet] (RNN3|-embed_layer.north) -- (RNN3);
    \draw[->,ultra thick,violet] (RNN3_2|-embed_layer.north) -- (RNN3_2);
    \draw[->,ultra thick,violet] (RNN4|-embed_layer.north) -- (RNN4);
    \draw[->,ultra thick,violet] (RNN4_2|-embed_layer.north) -- (RNN4_2);
    \draw[->,ultra thick,blue] (RNN5|-embed_layer.north) -- (RNN5);
    \draw[->,ultra thick,blue] (RNN6|-embed_layer.north) -- (RNN6);
    \draw[->,ultra thick,blue] (RNN7|-embed_layer.north) -- (RNN7);
    \draw[->,ultra thick,blue] (RNN8|-embed_layer.north) -- (RNN8);
    \node[yshift=-2em,text=violet] (input1) at (RNN1|-embed_layer.south) {8};
    \node[text=violet] (input2) at (RNN2|-input1) {29};
    \node[text=violet] (input3) at (RNN3|-input1) {2};
    \node[text=violet] (input3_2) at (RNN3_2|-input1) {17};
    \node[text=violet] (input_dots) at (RNN_dots|-input1) {$\cdots$};
    \node[text=violet] (input4) at (RNN4|-input1) {11};
    \node[text=violet] (input4_2) at (RNN4_2|-input1) {34};
    \node[text=blue] (input5) at (RNN5|-input1) {2};
    \node[text=blue] (input_dots2) at (RNN_dots2|-input1) {$\cdots$};
    \node[text=blue] (input6) at (RNN6|-input1) {8};
    \node[text=blue] (input7) at (RNN7|-input1) {11};
    \node[text=blue] (input8) at (RNN8|-input1) {29};
    \draw[->,ultra thick,violet] (input1) -- (RNN1|-embed_layer.south);
    \draw[->,ultra thick,violet] (input2) -- (RNN2|-embed_layer.south);
    \draw[->,ultra thick,violet] (input3) -- (RNN3|-embed_layer.south);
    \draw[->,ultra thick,violet] (input3_2) -- (RNN3_2|-embed_layer.south);
    \draw[->,ultra thick,violet] (input4) -- (RNN4|-embed_layer.south);
    \draw[->,ultra thick,violet] (input4_2) -- (RNN4_2|-embed_layer.south);
    \draw[->,ultra thick,blue] (input5) -- (RNN5|-embed_layer.south);
    \draw[->,ultra thick,blue] (input6) -- (RNN6|-embed_layer.south);
    \draw[->,ultra thick,blue] (input7) -- (RNN7|-embed_layer.south);
    \draw[->,ultra thick,blue] (input8) -- (RNN8|-embed_layer.south);
    \node[anchor=north west,align=center,text=violet] (input_ann) at (input1.south west) {\textbf{Study Items}\\(Non-Repeating Random Integers)};
    \path let \p1=(RNN5.west), \p2=(RNN8.east) in (RNN5) to (RNN8)
        node[above=2em of RNN5.north west, anchor=west, minimum width=\x2-\x1+0.5em, rounded rectangle, draw] (readout_layer) {Linear$\to$Softmax};
    \draw[->,ultra thick,orange] (RNN5) -- (RNN5|-readout_layer.south);
    \draw[->,ultra thick,orange] (RNN6) -- (RNN6|-readout_layer.south);
    \draw[->,ultra thick,orange] (RNN7) -- (RNN7|-readout_layer.south);
    \draw[->,ultra thick,orange] (RNN8) -- (RNN8|-readout_layer.south);
    \node[text=orange,yshift=2em] (sort5) at (RNN5|-readout_layer.north) {17};
    \node[text=orange] (sort_dots) at (RNN_dots2|-sort5) {$\cdots$};
    \node[text=orange] (sort6) at (RNN6|-sort5) {29};
    \node[text=orange] (sort7) at (RNN7|-sort5) {34};
    \node[text=orange] (sort8) at (RNN8|-sort5) {2};
    \draw[->,ultra thick,orange] (RNN5|-readout_layer.north) -- (sort5);
    \draw[->,ultra thick,orange] (RNN6|-readout_layer.north) -- (sort6);
    \draw[->,ultra thick,orange] (RNN7|-readout_layer.north) -- (sort7);
    \draw[->,ultra thick,orange] (RNN8|-readout_layer.north) -- (sort8);
    \node[anchor=south,text=orange] (sort_ann) at (sort7.north) {\textbf{Recall Successors}};
    \path let \p1=(input1.west), \p2=(input4.east) in (input1) to (input4)
        node[anchor=west, minimum width=\x2-\x1+1em, rounded rectangle, draw, dotted,thick,blue] (inputs) at (input1.west|-input4.east) {\phantom{M}};
    \path let \p1=(input5.west), \p2=(input8.east) in (input5) to (input8)
        node[anchor=west, minimum width=\x2-\x1+1em, rounded rectangle, draw, dotted,thick,blue] (shuffled) at (input5.west|-input8.east) {\phantom{M}};
    \draw[->,ultra thick,blue] (input4.south west) to [bend right=45] node [below,anchor=north west] {\textbf{Shuffle}} (input5.south east);
    \end{tikzpicture}